\documentclass{article} 
\usepackage{iclr2016_workshop,times}
\usepackage{amssymb,amsmath}
\usepackage{ifxetex,ifluatex}
\usepackage{fixltx2e} 

\ifnum 0\ifxetex 1\fi\ifluatex 1\fi=0 
  \usepackage[T1]{fontenc}
  \usepackage[utf8]{inputenc}
\else 
  \ifxetex
    \usepackage{mathspec}
    \usepackage{xltxtra,xunicode}
  \else
    \usepackage{fontspec}
  \fi
  \defaultfontfeatures{Mapping=tex-text,Scale=MatchLowercase}
  
\fi
\IfFileExists{upquote.sty}{\usepackage{upquote}}{}
\IfFileExists{microtype.sty}{%
\usepackage{microtype}
\UseMicrotypeSet[protrusion]{basicmath} 
}{}
\ifxetex
  \usepackage[setpagesize=false, 
              unicode=false, 
              xetex]{hyperref}
\else
  \usepackage[unicode=true]{hyperref}
\fi
\hypersetup{breaklinks=true,
            bookmarks=true,
            pdfauthor={},
            pdftitle={Understanding Visual Concepts with Continuation Learning},
            colorlinks=true,
            citecolor=blue,
            urlcolor=blue,
            linkcolor=magenta,
            pdfborder={0 0 0}}
\usepackage{graphicx,grffile}
\makeatletter
\def\maxwidth{\ifdim\Gin@nat@width>\linewidth\linewidth\else\Gin@nat@width\fi}
\def\maxheight{\ifdim\Gin@nat@height>\textheight\textheight\else\Gin@nat@height\fi}
\makeatother
\setkeys{Gin}{width=\maxwidth,height=\maxheight,keepaspectratio}
\title{Understanding Visual Concepts with Continuation Learning}
    \author{
                                        William F. Whitney,
                                                            Michael Chang,
                                                            Tejas Kulkarni,
                                                            and Joshua B. Tenenbaum\\
                Department of Brain and Cognitive Science\\
                Massachusetts Institute of Technology\\
                \texttt{\{wwhitney,mbchang,tejask,jbt\}@mit.edu}
                            }
\date{}

\usepackage{subfig}
\AtBeginDocument{%

}
\AtBeginDocument{%

}
\usepackage{float}
\floatstyle{ruled}
\makeatletter
\@ifundefined{c@chapter}{\newfloat{codelisting}{h}{lop}}{\newfloat{codelisting}{h}{lop}[chapter]}
\makeatother
\floatname{codelisting}{Listing}

\ifx\paragraph\undefined\else
\let\oldparagraph\paragraph
\renewcommand{\paragraph}[1]{\oldparagraph{#1}\mbox{}}
\fi
\ifx\subparagraph\undefined\else
\let\oldsubparagraph\subparagraph
\renewcommand{\subparagraph}[1]{\oldsubparagraph{#1}\mbox{}}
\fi

\begin{document}
\maketitle
\begin{abstract}
We introduce a neural network architecture and a learning algorithm to
produce factorized symbolic representations. We propose to learn these
concepts by observing consecutive frames, letting all the components of
the hidden representation except a small discrete set (gating units) be
predicted from the previous frame, and let the factors of variation in
the next frame be represented entirely by these discrete gated units
(corresponding to symbolic representations). We demonstrate the efficacy
of our approach on datasets of faces undergoing 3D transformations and
Atari 2600 games.
\end{abstract}

\section{Introduction}\label{introduction}

Deep learning has led to remarkable breakthroughs in solving perceptual
tasks such as object recognition, localization and segmentation using
large amounts of labeled data. However, the problem of learning abstract
representations of images without manual supervision is an open problem
in machine perception. Existing unsupervised learning techniques have
tried to address this problem (Hinton and Salakhutdinov
\protect\hyperlink{ref-hinton2006reducing}{2006}; Ranzato et al.
\protect\hyperlink{ref-ranzato2007unsupervised}{2007}; Lee et al.
\protect\hyperlink{ref-lee2009convolutional}{2009}) but lack the ability
to produce latent factors of variations or symbolic visual concepts from
raw data. Computer vision has historically been formulated as the
problem of producing symbolic descriptions of scenes from input images
(Horn \protect\hyperlink{ref-horn1986robot}{1986}). Without disentangled
and symbolic visual concepts, it is difficult to interpret or re-use
representations across tasks as no single component of the
representation vector has a semantic meaning by itself. Traditionally,
it has been difficult to adapt neural network architectures to learn
such representations from raw data. In this paper, we introduce a neural
network architecture and a learning algorithm to produce factorized
symbolic representations given consecutive images. We demonstrate the
efficacy of our approach on datasets including faces undergoing 3D
transformations, moving objects in 2D worlds, and Atari 2600 games.

\subsection{Related work}\label{related-work}

A number of generative models have been proposed in the literature to
learn abstract visual representations including RBM-based models (Hinton
and Salakhutdinov \protect\hyperlink{ref-hinton2006reducing}{2006}, Lee
et al. (\protect\hyperlink{ref-lee2009convolutional}{2009})),
variational auto-encoders (Kingma and Welling
\protect\hyperlink{ref-kingma2013auto}{2013}; Rezende, Mohamed, and
Wierstra \protect\hyperlink{ref-rezende2014stochastic}{2014}; Kulkarni
et al. \protect\hyperlink{ref-kulkarni2015deep}{2015}), convolution
based encoder-decoders (Ranzato et al.
\protect\hyperlink{ref-ranzato2007unsupervised}{2007}; Lee et al.
\protect\hyperlink{ref-lee2009convolutional}{2009}), and generative
adversarial networks (Goodfellow et al.
\protect\hyperlink{ref-goodfellow2014generative}{2014}; Radford, Metz,
and Chintala \protect\hyperlink{ref-radford2015unsupervised}{2015}).
However, the representations produced by most of these techniques are
entangled, without any notion of symbolic concepts. The exception to
this is more recent work by Hinton et al. (Hinton, Krizhevsky, and Wang
\protect\hyperlink{ref-hinton2011transforming}{2011}) on `transforming
auto-encoders' which use a domain-specific decoder with explicit visual
entities to reconstruct input images. Inverse graphics networks
(Kulkarni et al. \protect\hyperlink{ref-kulkarni2015deep}{2015}) have
also been shown to disentangle interpretable factors of variations,
albeit in a semi-supervised learning setting. Probabilistic program
induction has been recently applied for learning visual concepts in the
hand-written characters domain (Lake, Salakhutdinov, and Tenenbaum
\protect\hyperlink{ref-lake2015human}{2015}). However, this approach
requires the specification of primitives to build up the final
conceptual representations. The tasks we consider in this paper contain
great conceptual diversity and it is unclear if there exist a simple set
of primitives. Instead we propose to learn these concepts by observing
consecutive frames, letting all the components of the hidden
representation except a small discrete set (gating units) be predicted
from the previous frame, and let the factors of variation in the next
frame be represented entirely by these discrete gated units
(corresponding to symbolic representations).

\section{Model}\label{model}

\begin{figure}[htbp]
\centering
\includegraphics[width=1.00000\textwidth]{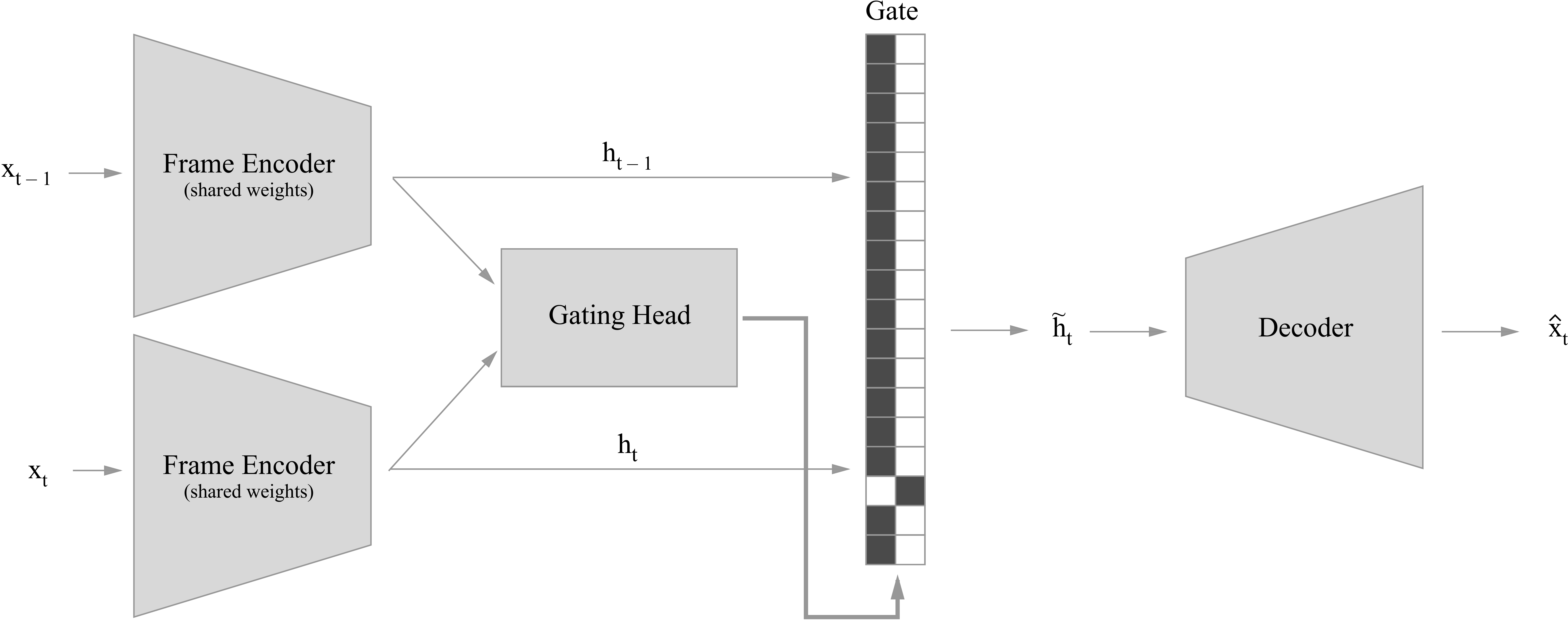}
\caption{\label{fig:model}The gated model. Each frame encoder produces a
representation from its input. The gating head examines both these
representations, then picks one component from the encoding of time
\(t\) to pass through the gate. All other components of the hidden
representation are from the encoding of time \(t-1\). As a result, each
frame encoder predicts what it can about the next frame and encodes the
``unpredictable'' parts of the frame into one component.}
\end{figure}

This model is a deep convolutional autoencoder (Hinton and Salakhutdinov
\protect\hyperlink{ref-hinton2006reducing}{2006}; Bengio
\protect\hyperlink{ref-bengio2009learning}{2009}; Masci et al.
\protect\hyperlink{ref-masci2011stacked}{2011}) with modifications to
accommodate multiple frames and encourage a particular factorization in
the latent space. Given two frames in sequence, \(x_{t-1}\) and
\(x_{t}\), the model first produces respective latent representations
\(h_{t-1}\) and \(h_t\) through a shared encoder. The model then
combines these two representations to produce a hidden representation
\(\tilde{h}_{t}\) that is fed as input to a decoder.

We train the model using a novel objective function: given the previous
frame \(x_{t-1}\) of a video and the current frame \(x_{t}\),
reconstruct the current frame as \(\hat{x}_{t}\).

To produce \(\tilde{h}_{t}\), we introduce a \emph{gating} in the
encoder (see Fig.~\ref{fig:model}) that select a small set of
\emph{gating units} that characterize the transformation between
\(x_{t-1}\) and \(x_t\). For clarity, in this paper we describe our
model under the context of one gating unit. Concretely, the encoder
learns to use a \emph{gating head} that selects one index \(i\) of the
latent representation vector as the gating unit, and then
\(\tilde{h}_{t}\) is constructed as \(h_{t-1}\), with the \(i\)th
component of \(h_{t-1}\) swapped out for the \(i\)th component of
\(h_t\).

Because the model must learn to reconstruct the current frame \(t\) from
a representation that is primarily composed of the components of the
representation of \(x_{t-1}\), the model is encouraged to represent the
attributes of \(t\) that are different from that of \(x_{t-1}\), such as
the lighting or pose of a face, in a very compact form that is
completely disentangled from the invariant parts of the scene, such as
the facial features. Thus, the model isolates the transformation from
\(x_{t-1}\) to \(x_{t}\) from other latent features via the component
\(i\) selected by the gating head.

\subsection{Continuation Learning}\label{continuation-learning}

To learn the gating function, we use a technique first described in (W.
Whitney \protect\hyperlink{ref-whitney2016disentangled}{2016}) for
smoothly annealing a soft weighting function into a binary decision.
Ordinarily, a model which produces a hard decision to gate through a
single component (out of e.g.~200) would be difficult to train; in this
case, it would require many forward passes through the decoder to
calculate the expectation of the loss for each of the possible
decisions. However, a model which uses a soft weighting over all the
components can be trained with gradient descent in a single
forward-backward pass.

In order to create a continuation between these two possibilities, we
use a scheduling for \emph{weight sharpening} (Graves, Wayne, and
Danihelka \protect\hyperlink{ref-graves2014neural}{2014}) combined with
noise on the output of the gating head. Given a weight distribution
\(w\) produced by the gating head and a sharpening parameter \(\gamma\)
which is proportional to the training epoch, we produce a sharpened and
noised weighting:

\[w_i' = \frac{\big(w_i + \mathcal{N}(0, \sigma^2)\big)^{\gamma}}{\sum_j w_j^{\gamma}}\]

This formulation forces the gating head to gradually concentrate more
and more mass on a single location at a time over the course of
training, and in practice results in fully binary gating distributions
by the end of training. This gating distribution thus selects a single
component of \(h_t\) to use in \(\tilde{h}_{t}\).

\section{Results}\label{results}

\begin{figure}[htbp]
\centering
\includegraphics[width=1.00000\textwidth]{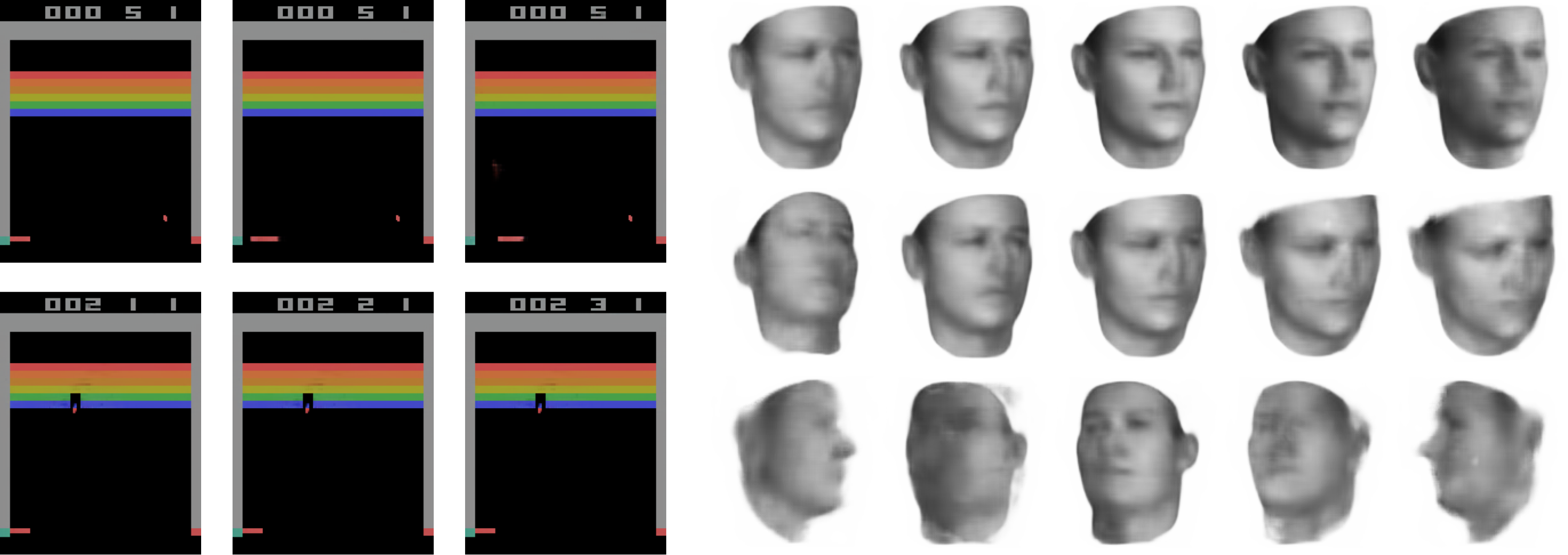}
\caption{\label{fig:results}\textbf{Manipulating the hidden
representation.} Each row was generated by encoding an input image, then
changing the value of a single component of the latent representation
before rendering it with the decoder. \textbf{Top left:} a single unit
controls the position of the paddle in Breakout. \textbf{Bottom left:}
another unit controls the count of the remaining lives in the score bar.
\textbf{Top right:} one unit controls the direction of lighting.
\textbf{Middle right:} a unit that controls the elevation of the face.
\textbf{Bottom right:} a unit controls the azimuth of the face, though
this transformation is not smooth. All input images are from the test
set.}
\end{figure}

\subsection{Atari frames}\label{atari-frames}

Our first dataset is frames from playing the Atari 2600 game Breakout.
The model is given as input two frames which occurred in sequence, then
reconstructs the second frame. This dataset was generated with a trained
DQN network (Mnih et al. \protect\hyperlink{ref-mnih2015human}{2015}).
Since the model can only use a few components of its representation from
the second frame, these components must contain all information
necessary to predict the second frame given the first. For this dataset
we use three gating heads, allowing three components of \(h_t\) to be
included in \(\tilde{h}_t\).

\subsection{Synthetic faces}\label{synthetic-faces}

We trained the model on faces generated from the Basel face model
(Paysan et al. \protect\hyperlink{ref-paysan2009face}{2009}) and
prepared as in (Kulkarni et al.
\protect\hyperlink{ref-kulkarni2015deep}{2015}). The input is two images
of the same face between which only one of \{lighting, elevation,
azimuth\} changes. For this dataset we use a single gating head, so the
model must represent all differences between these two images in one
unit only.

\section{Discussion}\label{discussion}

We have shown that it is possible to train a model which learns a
factorized, symbolic representation of the factors of variation in image
sequences from raw data. Such a model uses temporal continuity to
understand visual concepts at a high level, representing objects and
motion instead of raw pixels. Future work can extend this model to more
complex settings with an arbitrary number of factors of variation.

\section*{References}\label{bibliography}
\addcontentsline{toc}{section}{References}

\hypertarget{refs}{}
\hypertarget{ref-bengio2009learning}{}
Bengio, Yoshua. 2009. ``Learning Deep Architectures for AI.''
\emph{Foundations and Trends in Machine Learning} 2 (1). Now Publishers
Inc.: 1--127.

\hypertarget{ref-goodfellow2014generative}{}
Goodfellow, Ian, Jean Pouget-Abadie, Mehdi Mirza, Bing Xu, David
Warde-Farley, Sherjil Ozair, Aaron Courville, and Yoshua Bengio. 2014.
``Generative Adversarial Nets.'' In \emph{Advances in Neural Information
Processing Systems}, 2672--80.

\hypertarget{ref-graves2014neural}{}
Graves, Alex, Greg Wayne, and Ivo Danihelka. 2014. ``Neural Turing
Machines.'' \emph{ArXiv Preprint ArXiv:1410.5401}.

\hypertarget{ref-hinton2006reducing}{}
Hinton, Geoffrey E, and Ruslan R Salakhutdinov. 2006. ``Reducing the
Dimensionality of Data with Neural Networks.'' \emph{Science} 313
(5786). American Association for the Advancement of Science: 504--7.

\hypertarget{ref-hinton2011transforming}{}
Hinton, Geoffrey E, Alex Krizhevsky, and Sida D Wang. 2011.
``Transforming Auto-Encoders.'' In \emph{Artificial Neural Networks and
Machine Learning--ICANN 2011}, 44--51. Springer.

\hypertarget{ref-horn1986robot}{}
Horn, Berthold. 1986. \emph{Robot Vision}. MIT press.

\hypertarget{ref-kingma2013auto}{}
Kingma, Diederik P, and Max Welling. 2013. ``Auto-Encoding Variational
Bayes.'' \emph{ArXiv Preprint ArXiv:1312.6114}.

\hypertarget{ref-kulkarni2015deep}{}
Kulkarni, Tejas D, William F Whitney, Pushmeet Kohli, and Josh
Tenenbaum. 2015. ``Deep Convolutional Inverse Graphics Network.'' In
\emph{Advances in Neural Information Processing Systems}, 2530--8.

\hypertarget{ref-lake2015human}{}
Lake, Brenden M, Ruslan Salakhutdinov, and Joshua B Tenenbaum. 2015.
``Human-Level Concept Learning Through Probabilistic Program
Induction.'' \emph{Science} 350 (6266). American Association for the
Advancement of Science: 1332--8.

\hypertarget{ref-lee2009convolutional}{}
Lee, Honglak, Roger Grosse, Rajesh Ranganath, and Andrew Y Ng. 2009.
``Convolutional Deep Belief Networks for Scalable Unsupervised Learning
of Hierarchical Representations.'' In \emph{Proceedings of the 26th
Annual International Conference on Machine Learning}, 609--16. ACM.

\hypertarget{ref-masci2011stacked}{}
Masci, Jonathan, Ueli Meier, Dan Cireşan, and Jürgen Schmidhuber. 2011.
``Stacked Convolutional Auto-Encoders for Hierarchical Feature
Extraction.'' In \emph{Artificial Neural Networks and Machine
Learning--ICANN 2011}, 52--59. Springer.

\hypertarget{ref-mnih2015human}{}
Mnih, Volodymyr, Koray Kavukcuoglu, David Silver, Andrei A Rusu, Joel
Veness, Marc G Bellemare, Alex Graves, et al. 2015. ``Human-Level
Control Through Deep Reinforcement Learning.'' \emph{Nature} 518 (7540).
Nature Publishing Group: 529--33.

\hypertarget{ref-paysan2009face}{}
Paysan, P., R. Knothe, B. Amberg, S. Romdhani, and T. Vetter. 2009. ``A
3D Face Model for Pose and Illumination Invariant Face Recognition.''
\emph{Proceedings of the 6th IEEE International Conference on Advanced
Video and Signal Based Surveillance (AVSS) for Security, Safety and
Monitoring in Smart Environments}. Genova, Italy: IEEE.

\hypertarget{ref-radford2015unsupervised}{}
Radford, Alec, Luke Metz, and Soumith Chintala. 2015. ``Unsupervised
Representation Learning with Deep Convolutional Generative Adversarial
Networks.'' \emph{ArXiv Preprint ArXiv:1511.06434}.

\hypertarget{ref-ranzato2007unsupervised}{}
Ranzato, M, Fu Jie Huang, Y-L Boureau, and Yann LeCun. 2007.
``Unsupervised Learning of Invariant Feature Hierarchies with
Applications to Object Recognition.'' In \emph{Computer Vision and
Pattern Recognition, 2007. CVPR'07. IEEE Conference on}, 1--8. IEEE.

\hypertarget{ref-rezende2014stochastic}{}
Rezende, Danilo Jimenez, Shakir Mohamed, and Daan Wierstra. 2014.
``Stochastic Backpropagation and Approximate Inference in Deep
Generative Models.'' \emph{ArXiv Preprint ArXiv:1401.4082}.

\hypertarget{ref-whitney2016disentangled}{}
Whitney, William. 2016. ``Disentangled Representations in Neural
Models.'' \emph{ArXiv Preprint ArXiv:1602.02383}.

\end{document}